\documentclass{article}

\PassOptionsToPackage{dvipsnames}{xcolor}

\usepackage{microtype}
\usepackage{graphicx}
\usepackage{subfigure}
\usepackage{booktabs} 

\usepackage{hyperref}

\usepackage[accepted]{icml2021}

\usepackage[utf8]{inputenc} 
\usepackage[T1]{fontenc}    
\usepackage{nicefrac}       
\usepackage{epigraph}
\usepackage{enumitem}
\usepackage{todonotes}
\usepackage{paralist}
\usepackage{caption}

\usepackage{booktabs} 
\usepackage{dsfont}
\usepackage{multirow}
\usepackage{wrapfig}
\usepackage{float}
\usepackage{subfigure}

\usepackage{amsmath}
\usepackage{amssymb}
\usepackage{amsthm}
\usepackage{amsfonts}

\newcommand\dt{\Delta t}
\newcommand\dti{\Delta t_i}
\newcommand\dtiprev{\Delta t_{i-1}}
\newcommand\dtj{\Delta t_{j}}

\usepackage{color}
\definecolor{darkgreen}{rgb}{.1,0.4,.1}

\definecolor{darkblue}{rgb}{.1,0.1,.4}

\definecolor{darkred}{rgb}{0.6,0,0}

\icmltitlerunning{Piecewise-constant Neural ODEs} 

\begin{document}

\twocolumn[
\icmltitle{Piecewise-constant Neural ODEs} 

\icmlsetsymbol{equal}{*}

\begin{icmlauthorlist}
\icmlauthor{Sam Greydanus}{osu}
\icmlauthor{Stefan Lee}{osu}
\icmlauthor{Alan Fern}{osu}
\end{icmlauthorlist}

\icmlaffiliation{osu}{Oregon State University}
\icmlcorrespondingauthor{Sam Greydanus}{greydanus.17@gmail.com}

\icmlkeywords{RNNs, sequence models, neural networks, neural ODEs, autoregressive models}

\vskip 0.3in
]

\printAffiliationsAndNotice{}

\begin{abstract}
    Neural networks are a popular tool for modeling sequential data but they generally do not treat time as a continuous variable. Neural ODEs represent an important exception: they parameterize the time derivative of a hidden state with a neural network and then integrate over arbitrary amounts of time. But these parameterizations, which have arbitrary curvature, can be hard to integrate and thus train and evaluate. In this paper, we propose making a piecewise-constant approximation to Neural ODEs to mitigate these issues. Our model can be integrated exactly via Euler integration and can generate autoregressive samples in 3-20 times fewer steps than comparable RNN and ODE-RNN models. We evaluate our model on several synthetic physics tasks and a planning task inspired by the game of billiards. We find that it matches the performance of baseline approaches while requiring less time to train and evaluate.
\end{abstract}

\section{Introduction}

\begin{figure}[ht!]
    \centering
    \includegraphics[width=0.48\textwidth]{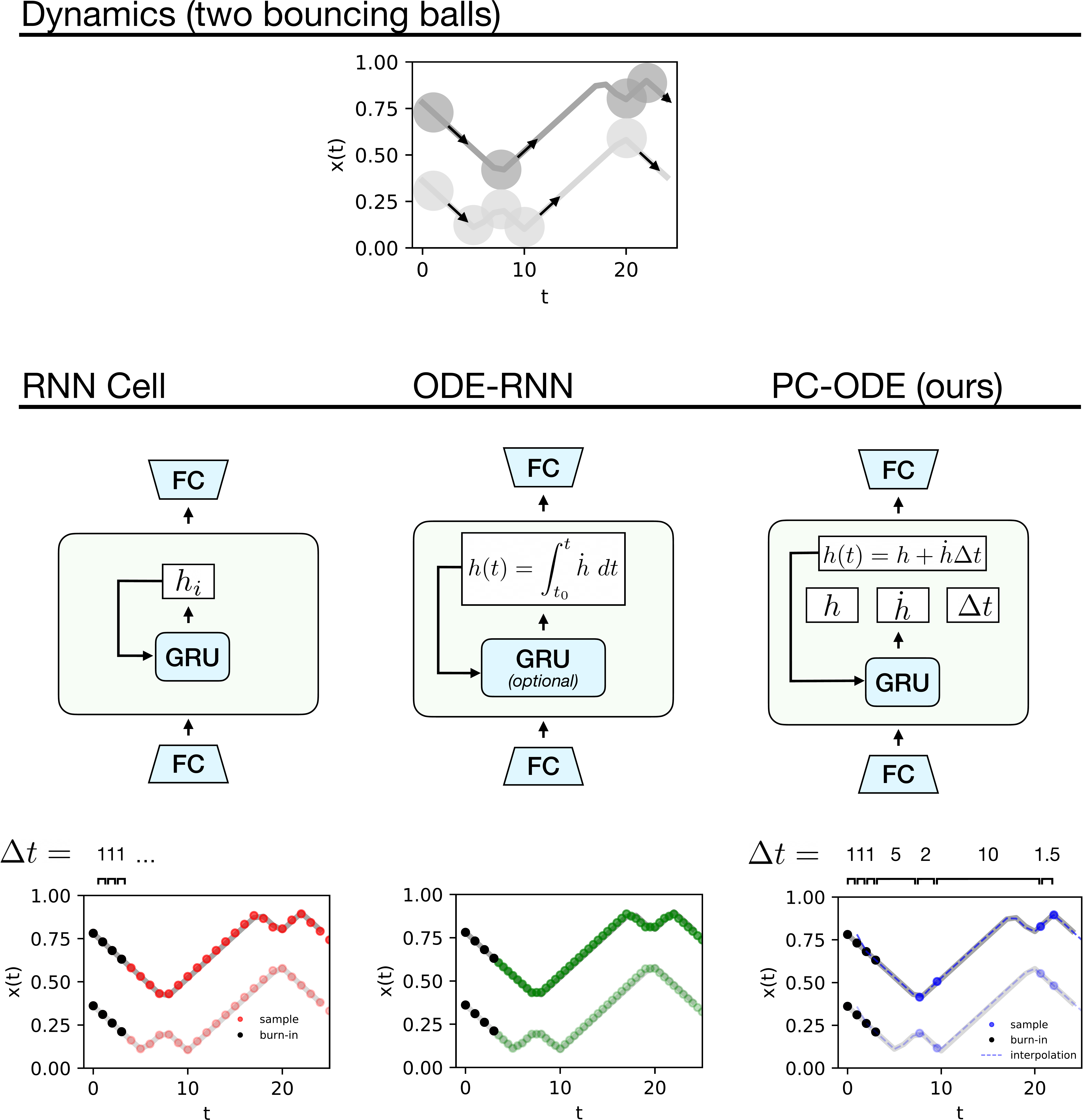}
    \caption{Predicting the dynamics of two billiards balls using an RNN (left), an ODE-RNN (center) and a PC-ODE (right). The RNN produces a hidden state $h_t$ at each time step and the ODE-RNN allows it to change continuously but is hard to integrate. Our model strikes a balance between the two by using a continuous-time hidden state that changes linearly over each interval $\dti$ and is thus easy to integrate. This allows it to skip over long spans of predictable motion and focus on key events such as collisions.}
    \label{fig:hero}
\end{figure}

It is said that change happens slowly and then all at once. Billiards balls move across a table before colliding and changing trajectories; water molecules cool slowly and then undergo a rapid phase transition into ice; and economic systems enjoy periods of stability interspersed with abrupt market downturns. That is to say, many time series exhibit periods of relatively homogeneous change divided by important events. Despite this, recurrent neural networks (RNNs), popular for time series modeling, treat time in uniform intervals -- potentially wasting prediction resources on long intervals of relatively constant change. 

A recent family of models called Neural ODEs \cite{chen2018neural} has attracted interest as a means of mitigating these problems. They parameterize the \textit{time derivative} of a hidden state with a neural network and then integrate it over arbitrary amounts of time. This allows them to treat time as a continuous variable. Integration can even be performed using adaptive integrators like Runge-Kutta \cite{runge1895numerische}, thus allocating more compute to difficult state transitions.

Adaptive integration is especially attractive in scenarios where ``key events'' are separated by variable amounts of time. In the game of billiards, these key events may consist of collisions between balls, walls, and pockets. Between these events, the balls simply undergo linear motion. That motion is not difficult to predict, but it is non-trivial for a model to learn to skip over it so as to focus on the more chaotic dynamics of collisions; this requires a model to employ some notion of \textit{temporal abstraction} \cite{sutton1999between, neitz2018adaptive}. This problem is not unique to billiards. The same challenge occurs in robotics, where a robot arm occasionally interacts with external objects at varying intervals. It may also occur in financial markets, scientific timeseries, and other environments where change happens at a variable rate.

One of the challenges of using adaptive integration to train neural networks is that the integrator must perform several function evaluations in order to estimate local curvature when performing an integration step. The curvature information determines how far the integrator can step forward in time, subject to a constant error budget. This is a particularly serious issue in the context of neural networks, which may have very irregular local curvatures at initialization. A single Neural ODE training step can take up to five times longer to evaluate than a comparable RNN architecture, making it challenging to scale these models \cite{rubanova2019latent}. The curvature problem has, in fact, already motivated some work on regularizing the curvature of Neural ODEs so as to train them more efficiently \cite{finlay2020train}. But even with regularization, these models are more difficult to train than RNNs. Furthermore, there are many tasks where regularizing curvature is counterproductive, for example, modeling elastic collisions between two bodies \cite{jia2019neural, chen2020learning}.

In this work, we propose a more direct approach to constraining curvature. We assume that our model's hidden state dynamics can be described by a sequence of piecewise-linear functions over variable temporal extents. Correspondingly, our Piecewise-constant ODE (PW-ODE) model defines the time derivative as a piece-wise constant function. This formulation enables trivial integration over arbitrary timespans. It also eliminates the need for complicated ODE solvers, since a simple Euler update of the form $h(\dt) = h_0 + \frac{dh_0}{dt}\dt$ is exact. Even though the resulting dynamics are piecewise-linear in latent space, they can represent highly nonlinear observation dynamics when paired with nonlinear encoders and decoders. These simplifications reduce training time, simplify implementation, and allow us to scale our model to non-linear time series with complex observations and chaotic dynamics.

We begin by describing the connection between Neural ODEs and our model. Then we perform a series of experiments, beginning with proof-of-concept tasks and proceeding to more difficult ones. At each stage, we compare our model to two baseline models: an ODE-RNN, which uses a Neural ODE to represent continuous hidden state dynamics, and a vanilla RNN, which has piecewise-constant hidden state dynamics. All code for this project is available online.\footnote{github.com/greydanus/piecewise\_node} Our core contributions are:

\begin{compactitem}
    \item Introducing a piece-wise constant Neural ODE that performs well and is simple to implement and train
    \item Evaluating it in comparison to ODE-RNNs and RNNs on several physics tasks, some of them chaotic and perceptually complex
    \item Showcasing its adaptive integration capabilities on a model-based planning task.
\end{compactitem}

\section{Piecewise-constant Neural ODEs}



\textbf{Neural ODEs.} Neural ODEs are a class of models that parameterize the time derivative of the hidden state using a neural network. Letting $\dot h(t) = f_{\theta}(x, t)$ where $f_{\theta}$ is the neural network and $x$ is an initial observation, we can solve for the dynamics of the hidden state by integration:
\begin{equation}
	h(t_1) = h(t_0) + \int_{t_0}^{t_1} \dot h(t) ~dt \label{eqn:neural_ode}
\end{equation}
This approach is rather general: we can use it to model any continuous-time dynamics we choose. But this same generality is also a weakness, for when we allow $f_\theta$ to have arbitrary, nonlinear curvature, it becomes difficult to integrate and train. In order to mitigate these issues, we would like to parameterize the hidden-state dynamics with a more constrained class of functions that are easier to integrate.

\textbf{RNN hidden state dynamics.} In contrast to Neural ODEs, RNNS entirely avoid integration by applying a potentially complex update function to the hidden state at fixed time intervals. So for RNNs, hidden state dynamics are piecewise constant over time. Letting $z(i)$ denote an encoding of an observation at time $i$, we can write these dynamics as:
\begin{equation}
    h(t) = h_i = RNN(h_{i-1}, z(i)) \mbox{~~for~~} t\in [i, i{+}1)
\end{equation}
While simple to evaluate, piecewise-constant dynamics prevent RNN hidden states from changing between cell updates. This makes selecting an appropriate step size an important hyperparameter. Too large and fine detail will be lost. Too small and commputational costs increase for long time spans. Regardless, the step sizes must be of uniform duration.

Recent works including ``Neural Event Functions'' in \citet{chen2020learning} and ODE-RNNs in \citet{rubanova2019latent} have sought to make RNNs more expressive by replacing these piecewise-constant dynamics with Neural ODEs between time steps. However, these models still must contend with complex and expensive integration of arbitrary curvature.

\textbf{Piecewise-constant ODEs (PC-ODEs).} We follow a similar strategy to these works, but restrict ourselves to a simple model class where integration between time steps can be computed efficiently. We make a locally-linear approximation of hidden-state dynamics, writing the Taylor series expansion around $h(t_0)$, 
\begin{align}
	h(t) &= h(t_0) + \dot h(t_0)(t-t_0)+\tfrac{1}{2} \ddot h(t_0)(t-t_0)^{2}\dots \label{eqn:taylor}\\
	&\approx h(t_0) + \dot h(t_0)\dt
\end{align}
This simple, first-order approximation assumes that changes in the hidden state $h$ are close to linear with respect to $\dt$. Unlike Neural ODEs with arbitrary curvature, integrating this linear form is trivial; a step of Euler integration is exact.

While local-linearity may seem like a strong assumption, we note that this is with respect to the \emph{hidden state} and joint training with complex, non-linear encoders and decoders can learn mappings that are linear to achieve low loss. This may hold even in cases where the observation space is perceptually complex. Consider, for example, a video of a ball undergoing linear motion. While the dynamics are simple when projected into the right coordinate system, they vary in a nonlinear way in pixel space. Sufficiently complex encoders and decoders may learn this transformation if constrained to linear dynamics within the hidden state.

The error induced by this locally-linear approximation depends on the curvature of the underlying function. Analogously to adaptive integration methods, a good piecewise-linear approximation should produce many short linear segments in areas of high curvature and longer ones when change in the hidden state is more constant.

To formalize this setup, we consider a sequence of non-uniform time intervals $\Delta t_i$ with timestamps $\tau_i$. Put another way, $\tau_i = \sum_{j=0}^{i-1} \dtj$. This lets us define our hidden state as:
\begin{eqnarray}
	h(t) =& h(\tau_i) + \dot h(\tau_i) (t-\tau_i)
	\label{eqn:euler}
\end{eqnarray}
where the index $i$ above denotes the most recent timestep in the sequence. This model allows $h$ to change in a continous linear fashion between timestamps. And at each timestamp $\tau_i$, $h$ can undergo arbitrary discontinuous updates (computed by an RNN cell update).

\textbf{Implementing PC-ODEs as RNNs.} 
To implement PC-ODEs, we need to predict a sequence of locally-constant ODEs, parameterized by  $h(\tau_i)$, $\dot h(\tau_i)$, and the duration of each segment, $\dti$. 
As $h(\tau_i)$ represents a discontinuous change in $h$ at $\tau_i$, we predict it directly using the RNN cell update.

We build on a standard RNN cell -- predicting $h_{\tau_i}$, $\dot h_{\tau_i}$, and $\dti$ at each cell update. Letting $[]$ denote concatenation and $z(\tau_i)$ the encoding of an observation at time $\tau_i$, we write
\begin{eqnarray}
	h_{\tau_i}, \dot h_{\tau_i} =& \mbox{RNNCell}\left( \left[h_{\tau_{i-1}}, \dot h_{\tau_{i-1}}\dtiprev \right],  z(\tau_i) \right) \\
	\dti =& 1 + \mbox{LeakyReLU}(Wh_{\tau_i}+b) \label{eq:leakyrelu} 
\end{eqnarray}
To compute the hidden state at some time $t$, we simply use Equation \ref{eqn:euler}.
This hidden state can be decoded to an output $\hat{x}(t)$ by a decoder. Note that this system has two notions of time -- continuous time denoted $t$ and a sequence $i=0,1,...$ of RNN cell updates occurring at continuous times $\tau_i$ depending on predicted durations $\dti$. 
In contrast to a standard RNN, which operates at a fixed time intervals, our underlying RNN model jumps forward in time dynamically. In Equation \ref{eq:leakyrelu}, we bias our model to always stepping forward at least one unit of time.

One challenge associated with this model is that the optimal value of $\dti$ changes depending on the predicted dynamics. The dynamics, in turn, depend on all previous values of $\dt$. Given that these variables are in a dynamic relationship, we cannot use standard teacher forcing -- where $\dt$ is always one -- for training. Rather, we must use a variant of the approach which dynamically solves for the optimal value of $\dt$. We will discuss how to do this in the next section.

\section{Training our model} \label{sec:training_details}

Even though our model is defined over continuous time, training data for time-series models typically consists of observation sequences of the form $x_0, x_1, x_2, \dots, x_T$. For the purposes of this work, we will assume that these observations are separated by a constant $\dt=1$.

When training a standard RNN on a sequence modeling task, standard practice is to directly minimize the error of producing the next output in the ground truth sequence given the preceding ones. This loss can be written $\mathcal{L}_x = \sum_{t=0}^{T} \ell(x_t, \hat{x}(t))$, where $\ell(\cdot, \cdot)$ is a loss between predicted and actual values. However, in order to learn an adaptive timestepping behavior, we must also encourage our model to take large time steps  (predict large $\dti$) when possible.

One way to strike a balance between accuracy and computational efficiency is to limit the amount of error associated with each step. This is analogous to setting the error tolerance of an ODE solver, but here the size of the time step is determined by $\dti$. Let us define this requirement as $\ell(x_t, \hat{x}(t)) < \epsilon$ where $\epsilon$ is a hyperparameter that determines the model's error tolerance.\footnote{We recognize the importance of minimizing the number of hyperparameters in a model. This hyperparameter is inevitable, though, as one cannot perform adaptive integration without setting an error tolerance.} Now our goal is to predict $\dti$ to be as large as possible while satisfying this requirement. In other words, we seek to find $\dti^*$ which satisfies
\begin{eqnarray}
     \max_{\dti \geq 1}~~ \dti ~~~~ \mbox{s.t.} ~~~~~~ \ell\left(x_t, \hat{x}(t)\right) < \epsilon \label{eq:linesearch}\\\nonumber
     \forall \left\{x_t \mid t \in [\tau_i, \tau_i+\dti]\right\}.
\end{eqnarray}
This can be solved using a simple forward line search that starts at $\tau_i$ and moves forward through the samples until our model's prediction loss under the current dynamics exceeds $\epsilon$. If this occurs on the first step (implying no time passes between ticks), we set $\dti^*$ to 1. In Appendix \ref{appendix:batchwise_search} we show this can be efficiently vectorized for an entire batch as part of a single teacher forcing pass.\footnote{This is a practical improvement over Neural ODEs, which are not easy to minibatch and integrate adaptively; see discussions in \citet{chen2018neural} and \citet{rubanova2019latent}.} We compute the optimal $\dti^*$ on-the-fly during training and step forward accordingly -- effectively ignoring the predicted $\dti$, as is also done in standard teacher forcing training. To update the model to predict the correct $\dti^*$, we augment the prediction loss $L_x$ with a step size loss $\mathcal{L}_{\dt} = \sum_{i} ||\dti - \dti^*||_2^2$. At inference, the predicted $\dti$ sets the step size.

\textbf{Interpreting the error threshold.} Examining Equation \ref{eq:linesearch}, $\epsilon$ acts as a trade-off between step size and approximation error. In the extreme of setting $\epsilon=0$, the model predictions will likely never be below $\epsilon$, $\dti$ will remain at 1, and our model will perform cell updates at the same rate as a standard RNN. At the other extreme, choosing $\epsilon=\infty$ forces our model to predict entire time series with a constant ODE. In practice, we found that setting $\epsilon$ to the final training loss of the baseline RNN model\footnote{We use the baseline RNN model because it performs better than the baseline Neural ODE model, thus offering a more challenging target for prediction accuracy.} yields models that update 3x-20x less than a standard RNN while maintaining the same test error.

\section{Experiments}

\begin{figure*}[t]
\centering
\includegraphics[width=\textwidth]{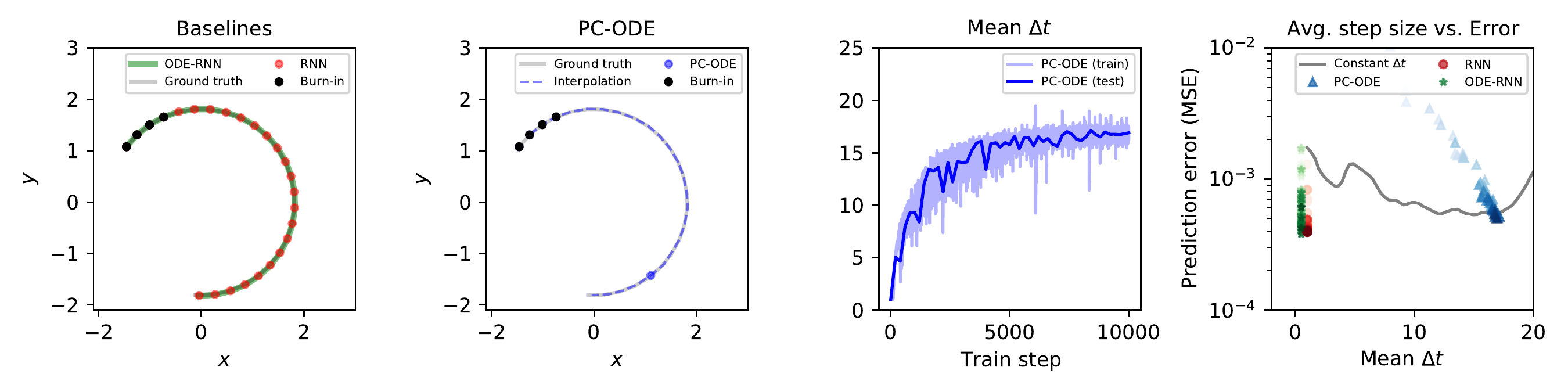}
\caption{A comparison between PC-ODE and two baseline models trained the circles task. \underline{Columns 1 and 2} show samples from the models. The PC-ODE model summarizes almost the entire trajectory with a single step.  \underline{Column 3} shows how step sizes increase throughout training. \underline{Column 4} shows that our model is able to produce autoregressive samples of comparable quality to the baseline while using twenty times fewer cell updates.}
\label{fig:circles_summary}
\end{figure*}

We evaluate our model on a suite of tasks including both coordinate and video representations of a small-scale billiards simulation. Our model attains test errors and autoregressive sample errors comparable to those of baseline models while performing 3-20x fewer steps. We also show how to leverage our model for model-based planning.

\textbf{Implementation Details.} We use the same architecture and training regime across all experiments unless otherwise specified. We use a GRU cell with $128$ hidden units and our encoder and decoder networks are four-layer residual MLPs with $128$ hidden units and ReLU activations. We train our model with the Adam optimizer for $10,000$ steps with a batch size of $256$, a learning rate of $1 \times 10^{-3}$ and a decay schedule of $\gamma=0.9$ every $5000$ steps. We used a mean square error loss function to train both teacher forcing and $\dt$ predictions, and scaled the latter by $1 \times 10^{-5}$ relative to the former. For model selection, we used early stopping.

The initial transition to stepping more than one step is challenging as it requires latent dynamics to extrapolate essentially without supervision to do so. To bootstrap, we force a step larger than one even when error is greater than $\epsilon$ in 1\% of cases. This increases stability; for more discussion see Appendix \ref{appendix:bootstrapping}.

\textbf{Baselines.} As baselines, we consider a vanilla RNN trained at a fixed time step ($\dt=1$) and an ODE-RNN \cite{rubanova2019latent} which also includes an RNN operating at a fixed time step but predicts the parameters of an ODE at each step in order to integrate continuous hidden states between steps. The same encoders and decoders are used for all models. In early experiments, we tried using a Neural ODE to solve for hidden state dynamics without intervening RNN cell updates; such a model might be called a latent Neural ODE. In nearly all experiments, it was worse than the ODE-RNN by at least an order of magnitude.

\subsection{Predicting linear motion}

To begin, we consider two simple settings that involve an object undergoing constant linear motion. These settings demonstrate the limitations of standard RNNs and ODEs and showcase our model's ability to translate linear latent dynamics into non-linear motion. The first of these settings is the Lines task, which involves motion along a line.

We generate 10,000 trajectories corresponding to linear motion along one dimension in $(x,y)$ Cartesian coordinates. Specifically, $\{(x_0, y_0), \dots , (x_{20}, y_{20})\}$ where $x_t = t$ and $y_t = c$ where $c$ is a random constant in $(0,1)$. 
This corresponds to a flat, straight line starting at a random height. Unsurprisingly, our model quickly learns to jump to the end of the sequence (mean $\dti$ of 20) while the RNN operates at a fixed rate. Although it is theoretically possible for the ODE-RNN to summarize this trajectory in one integration step, in practice it required more function evaluations (and walltime) to train than the RNN. We trained on $90\%$ of the data and evaluated on the remainder and all models trivially achieved very low error ($10^{-8}$). See Appendix \ref{appendix:visualize_lines} for samples.

\subsection{Predicting circular motion}

Our second task also involves constant motion, but here the motion is circular. Even though this motion is non-linear in the Cartesian observation space, it \textit{is} linear in polar coordinates. This experiment examines our model's ability to encode non-linear observations with linear latent dynamics -- effectively learning a Cartesian-to-Polar conversion. As before, our dataset contained $10,000$ examples, this time with $25$ time steps each. Each sequence started from a random angle in $(0,2\pi)$ and a random radius in $(1,2)$. We fixed the tangential velocity to a constant value across all trajectories based on the intuition of a particle undergoing constant velocity along 1D manifolds of various curvatures.

As in the lines task, our model learned to step through significant spans of time (mean $\dti$ of 17). Figure \ref{fig:circles_summary} provides our primary visualization of these results. The first two columns show representative samples from the three models. Dots correspond to RNN updates with black dots corresponding to priming observations. Recall that our model can make a prediction at any time point between updates. We plot these intermediate evaluations with a dashed line and they follow the ground truth trajectory closely. The third column shows the average optimal step size $\dti^*$ during training. As we can see, it rises during training as the model learns.

The fourth column presents a key summary plot of our experiment -- a comparison of mean squared error vs. efficiency in terms of step size. In this chart, further lower-right is better, and corresponds to low error with large step sizes. In green and red, we show the progression of error for the ODE-RNN and RNN over training (light to dark). Likewise, we show our model's progress in blue. Unlike the RNN which is at a fixed $\dt$ or the ODE-RNN, which performs even more function evaluations than the RNN, our approach changes both in error and time step size. We see that our approach is significantly jumpier while maintaining a similar error rate.

The gray curve represents the performance of the our model when it is forced to take constant step sizes of size $\dt$. In this task, our model converges to a point only slightly beyond this curve, indicating limited utility in dynamic $\dti$ prediction compared to the optimal fixed time step. This is unsurprising because the dynamics are uniform throughout each sequence. We will see in subsequent experiments that dynamic predictions of $\dti$ can be impactful in more complex settings. Finally, in Figure \ref{fig:capacity}, we vary the dimensionality of our model's hidden state and retrain; we find that model capacity and step size are positively correlated.

\begin{center}
    \includegraphics[clip=true, trim=0.25cm 0.55cm 0.1cm 0.2cm, width=0.4\textwidth]{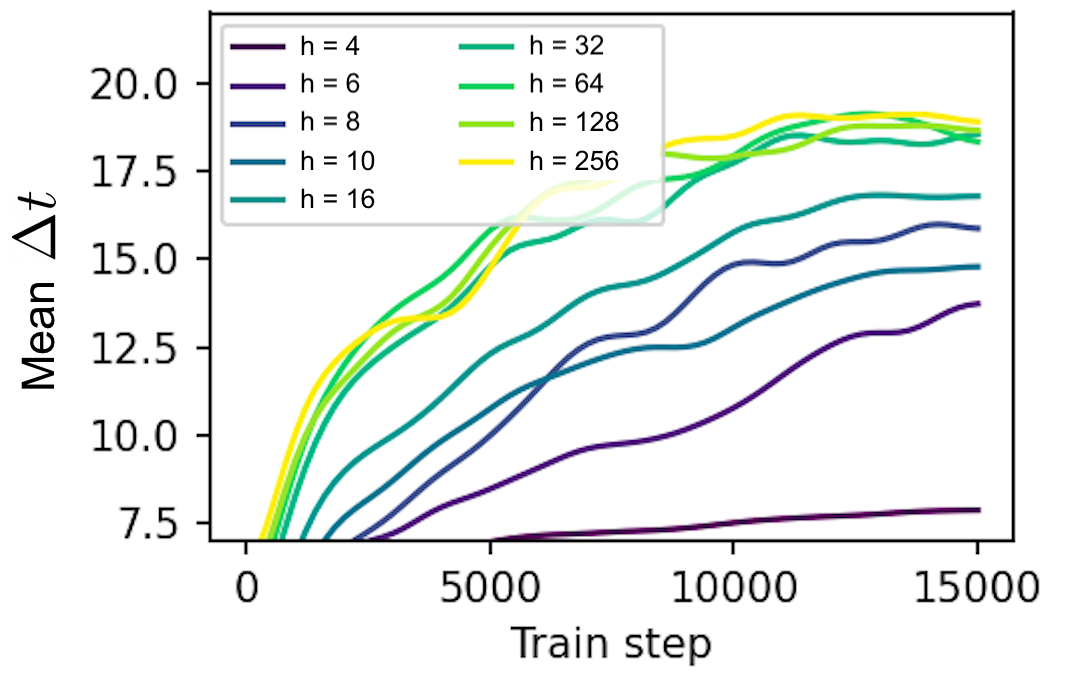}
    \captionof{figure}{Mean step size increases as we increase model capacity (hidden state size). We observed this effect across all datasets and experiments.}
    \label{fig:capacity}
\end{center}

\begin{figure*}
    \begin{center}
      \includegraphics[width=\textwidth]{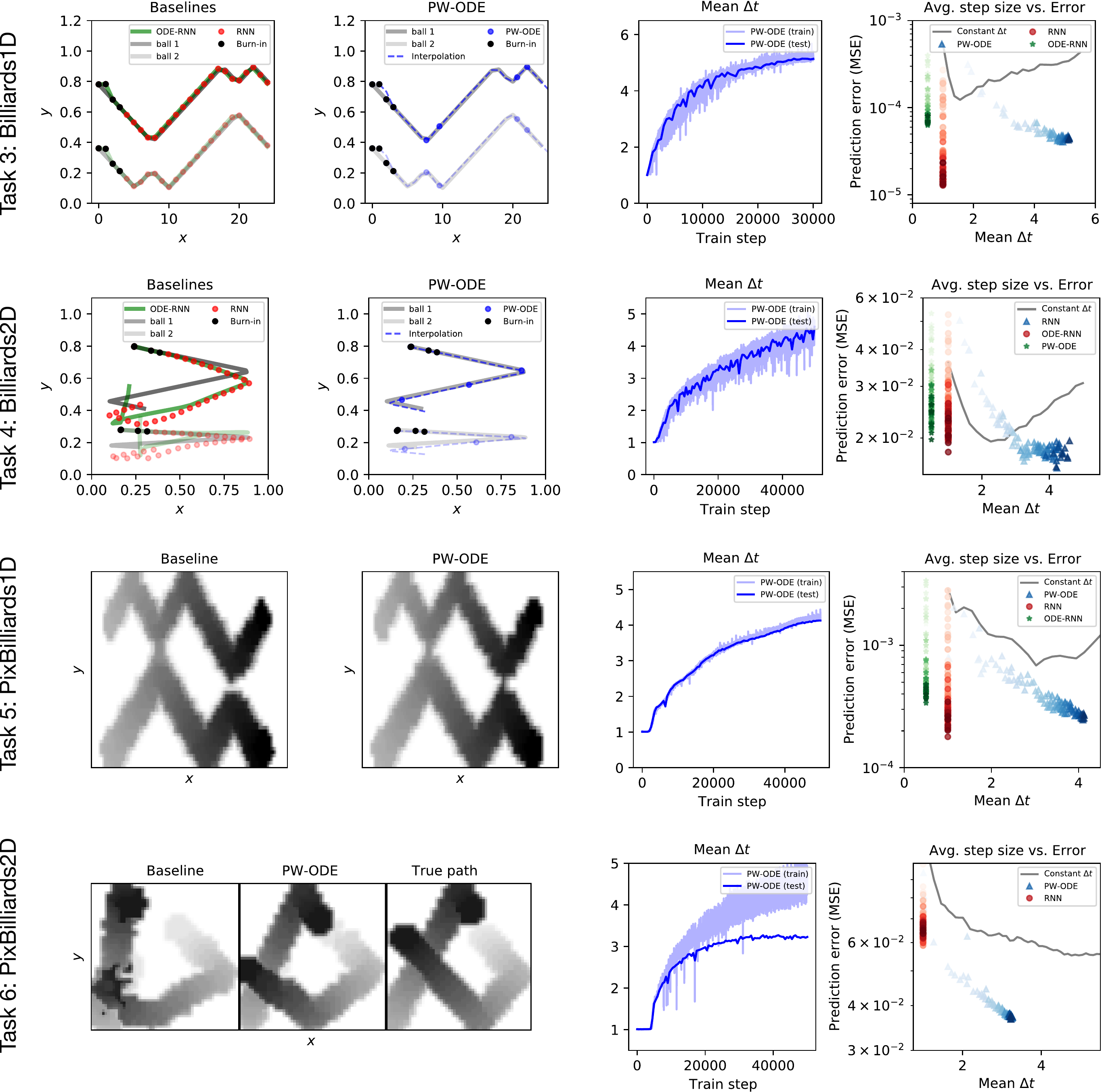}
        \captionof{figure}{Comparing PC-ODE to baseline models trained on four billiards tasks. \underline{Columns 1 and 2} illustrate qualitatively that our model's predictions are at least as good as those of the baselines. RNN samples are shown for the pixel tasks, as they improve over the ODE-RNN. \underline{Columns 3 and 4} illustrate that our our model's time stepping varied adaptively, giving larger time steps on easy tasks and shorter time steps on more difficult ones. Meanwhile, it matched or improved over the predictions of the baseline models while being more efficient to sample from. Forcing our model to use a constant $\dt$ (gray line) gives much worse performance. This suggests that our model's \textit{adaptive time stepping} skips over the most predictable time spans and concentrates on the less predictable ones. \underline{Supplementary Materials} contains videos of predicted trajectories.}
    \label{fig:summaries_p2}
    \end{center}
\end{figure*}

\begin{table*}[t!]
    \setlength{\tabcolsep}{9pt}
    \begin{tabular}{lccccccc}
    \toprule
         & & \multicolumn{2}{c}{Test MSE} & \multicolumn{2}{c}{Sample MSE} & \multicolumn{2}{c}{Mean $\dti$} \\
         \cmidrule(lr){3-4}\cmidrule(lr){5-6}\cmidrule(lr){7-8}
    Task & MSE Scale & Baselines & PC-ODE & Baselines & PC-ODE & Baselines & PC-ODE \\ \midrule
    1: Lines & $\times 10^{-8}$ & $1.77/1.77$ & $44.2$ & $17.9/\mathbf{13.9}$ & $21.9$ & $1/1$ & $\mathbf{20}$ \\
    2: Circles & $\times 10^{-4}$ & $0.03/0.03$ & $3.64$ & $5.47/\mathbf{3.98}$ & $5.25$ & $1/1$ & $\mathbf{17}$ \\
    3: Billiards1D & $\times 10^{-5}$ & $2.44/2.43$ & $2.93$ & $7.29/\mathbf{1.77}$ & $4.37$ & $1/1$ & $\mathbf{5.1}$ \\
    4: Billiards2D & $\times 10^{-5}$ & $5.13/2.19$ & $2.37$ & $2700/2360$ & $\mathbf{1570}$ & $1/1$ & $\mathbf{2.71}$ \\
    5: PixBill1D & $\times 10^{-4}$ & $1.57/1.57$ & $1.74$ & $3.87/\mathbf{2.37}$ & $2.64$ & $1/1$ & $\mathbf{4.1}$ \\
    6: PixBill2D & $\times 10^{-3}$ & $1.99/2.08$ & $3.90$ & $64.1/64.8$ & $\mathbf{37.1}$ & $1/1$ & $\mathbf{3.2}$ \\
    \bottomrule
    \end{tabular}
    \caption{Quantitative comparison between PC-ODE and baseline models, denoted here as Baselines collectively for brevity. We present statistics from the two baseline models in the format ODE-RNN/RNN.}
    \label{tab:quant}
\end{table*}

\subsection{Billiards from coordinates}

To examine more complex dynamics, we consider a billiards-like simulation in which two balls bounce around within a walled enclosure. We examine this setting for 1- and 2-dimensional versions from both coordinate and image-based inputs. Ball trajectories in this setting exhibit periods of essentially linear motion punctuated by collisions. As such, we expect our PC-ODE model to adjust its step sizes so as to have cell updates coincide with collision events.

In each setting, we generate $10,000$ trajectories of 45 time steps each. The environment lacks friction, collisions are perfectly elastic, and walls are placed along 0 and 1 for each coordinate. For coordinate settings, we record trajectories as $x,y$ coordinates for each ball (or just $y$ in the 1D case). For image settings, we render $28\times28$ grayscale images depicting a top-down view of the balls. 

The first two rows of Figure \ref{fig:summaries_p2} visualize the result of 1D and 2D billiards respectively. To visualize samples in the 1D case, we treat time as the x-axis. For 2D, we fade from light to dark to indicate time. In samples from both cases, we see the model learns to concentrate cell updates near collisions. However, we do see instances where collisions are skipped entirely and filled in with linear latent interpolation alone as in the 1D example. The `Step size vs.~error' plots (last column) also reflect the impact of this adaptive timestepping. Our approach is able to improve in both accuracy and average step size compared to using a fixed step size (darkest blue triangle vs.~gray curve). For 1D, the baseline models achieve lower error but require roughly 6x the compute. In the 2D case, our model outperforms the baselines while executing 4x fewer updates. 

\subsection{Billiards from pixels}

Learning from pixels directly is a challenging setting. While the underlying motion of the balls is the same as before, the way pixels change as the balls move is extremely non-linear as individual pixels turn on or off abruptly. For these experiments, we increase the hidden layer size to 512 units and train for $50,000$ steps. The third and forth rows of Figure \ref{fig:summaries_p2} show result summaries for 1D and 2D pixel-based billiards. In the 1D setting, we find that both the baselines and our model achieve similar error, but our model does so with 4x fewer steps. In the 2D setting however, the baseline RNN yields a 50\% higher error rate and generates unrealistic trajectories where balls bend along their trajectory like the sample shown. The ODE-RNN is even worse. Reviewing training curves, we noted that both baseline models overfit more severely.

\textbf{Quantitative results.} All experimental settings are summarized with dataset-level metrics at convergence in Table \ref{tab:quant}. We find our PC-ODE model performs at or near the same sample error rate while making 3 to 20 times fewer RNN steps. Note that absolute scale of error terms varies by orders of magnitude across tasks and is specified only in the \texttt{MSE Scale} column for compactness.

\subsection{Model-based Planning} 

Our final experiment demonstrates our PW-ODE model in a simple model-based planning application in the 2D billiards domain. We consider a simplified billiards game consisting of a $1\times1$ table with a single corner pocket at (1,0) and two balls -- a cue ball and a target ball. Both balls start at rest. The planning task is to set initial $x,y$ velocities of the cue ball (i.e.~take a shot) such that the target ball enters the pocket (< 0.17 units from (1,0)) within 35 time steps. As our goal is to examine the usefulness of our learned dynamics models (and not to make an advance in planning), we use a simple random search procedure.

We consider 250 random initial configurations of the balls. For each, we generate random initial $x,y$ velocities for the cue ball as candidate actions. For each candidate, we simulate the resulting trajectory using each of our learned models. We select the action that results in the minimum distance between the target ball and the pocket over the course of the rollout. We evalate this action by executing it in the ground-truth simulator.

\begin{figure}[t]
\begin{center}
    \includegraphics[clip=true, trim=0.25cm 0.35cm 0.1cm 0.3cm, width=0.425\textwidth]{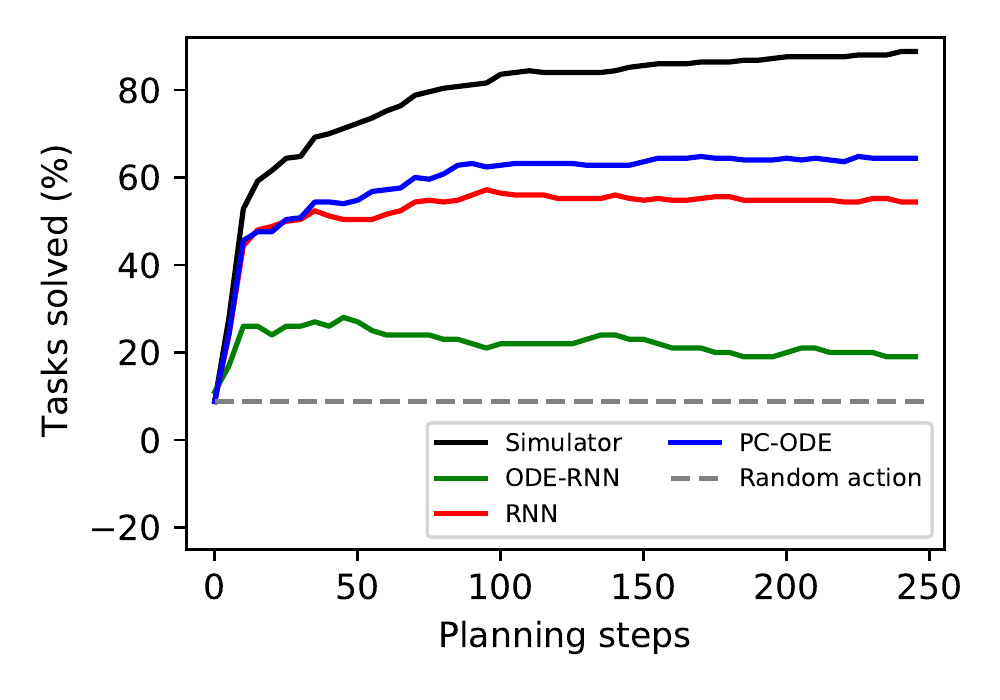}
    \caption{We compare PC-ODE to our two baselines (RNN and ODE-RNN) and the ground-truth simulator on a billiards planning task. We find our model is comparable to an RNN-based planner and significantly better than the ODE-RNN planner. Further, our model requires 3x fewer function evaluations than the RNN and >6x fewer than ODE-RNN.}
    \label{fig:planning}
\end{center}
\end{figure}

Figure \ref{fig:planning} shows the success rates of our models as a function of the number of planning rollouts they perform. We compare our model with the baselines, an upper bound using the actual simulator to rank actions, and a random action lower bound.  We find our PC-ODE model outperforms the other learned methods -- achieving a 61\% success rate. The RNN performs nearly as well but executes almost four times more cell updates than our model. The ODE-RNN performs significantly worse because of its poor predictive accuracy. The wallclock time of the ODE-RNN was also far worse, 3 times worse than the others, especially when evaluated without a GPU.

\section{Related Work}

\textbf{Neural ODEs.} The concept of neural ODEs has existed since the  1990's \citep{lecun1988theoretical, pearlmutter1995gradient}, but has recently seen a resurgence of interest due to \citep{chen2018neural}. As already discussed, while these models and their piecewise variants \citet{rubanova2019latent,chen2020learning} have shown promise, computational efficiency is still a challenge. Our work addresses this by imposing a restricted model that allows for trivially efficient integration. An important point of future work is to better understand if and when the additional computational overhead of general neural ODEs pays off in practice, especially when coupled with powerful encoders and decoders.

\textbf{RNNs for Irregular Timeseries.} RNN models have been studied for handling time series with irregular or missing time steps, e.g. including the time gap between observations as an input to the RNN cell. In practice, though, preprocessing the data by interpolating \cite{che2018recurrent} or masking \citep{lipton2016directly} and then training a regular RNN tends to work better. Several works have proposed applying an exponential decay function to the RNN hidden state in proportion to the duration of time between observations \citep{che2018recurrent,cao2018brits,rajkomar2018scalable, mozer2017discrete}. This approach makes hidden state dynamics continuous in time in a limited sense: hidden states are constrained to decay towards zero at a particular rate. In contrast, our approach for adaptively predicting piecewise linear dynamics of arbitrary length offers more flexibility.

\textbf{Related RL and Planning Models.} Recent work by \citet{gregor2018temporal} in the context of reinforcement learning develops a jumpy planning model which does not use an RNN cell or perform continuous interpolation of latent states. Another relevant work is Embed to Control by \citet{watter2015embed} which, like our model, assumes that dynamics is linear in a latent space. As with our work, they train their model on several dynamics problems, some of them from pixels. Unlike our work, their model performs inference over uniform time steps and does not learn an adaptive timestepping behavior.

Related work such as Adaptive Skip Intervals \citep{neitz2018adaptive} and Time-Agnostic Prediction \citep{jayaraman2018time} use a ``minimum-over-time'' loss across a horizon of future states. This allows them to predict the next predictable state, regardless of how distant in time it might be to support temporally-abstract planning.  However, they cannot reconstruct intervening states, estimate the ``rate of change'' of the system, or treat time as a continuous quantity.  Infobot \citep{goyal2019infobot} includes a similar notion of temporal abstraction but parts of Infobot still tick at every time step and the model treats time as a discrete variable.

\textbf{Temporal Abstraction in RNNs.}  \citet{koutnik2014clockwork} proposed dividing an RNN internal state into groups and only performing cell updates on the $i^{th}$ group after $2^{i-1}$ time steps. More recent works have aimed to make this hierarchical structure more adaptive, either by data-specific rules \citep{ling2015character} or by a learning  mechanism \citep{chung2019hierarchical}. More recently, \cite{goyal2019recurrent} have proposed Recurrent Independent Mechanisms, which can learn to make dynamics predictions semi-independently at different time scales. All these models exhibit temporal abstraction, but parts of them -- the encoders at the very least -- must update at each time step. Additionally, they do not treat time as a continuous variable.

\section{Conclusions}

Neural networks are already a widely used tool, but they still have fundamental limitations. In this paper, we reckoned with the fact that they struggle at adaptive timestepping and the computational expense of integration. In order to make RNNs and Neural ODEs more useful in more contexts, it is essential to find solutions to such restrictions. With this in mind, we proposed a PC-ODE model which can skip over long durations of comparatively homogeneous change and focus on pivotal events as the need arises. We hope that this line of work will lead to models that can represent time more efficiently and flexibly.


\bibliographystyle{abbrvnat}
\setlength{\bibsep}{5pt} 
\setlength{\bibhang}{0pt}
\bibliography{references}

\newpage
\appendix
\section{Additional Methods}
\subsection{Efficient Batch-wise Line Search with Masking.} \label{appendix:batchwise_search}

In Section \ref{sec:training_details} we defined the optimal jump width, $\dti^*$, and described how to find it for a given step $i$ using the constrained optimization objective in Equation \ref{eq:linesearch}. Then we described how one might solve that equation by starting from $\tau_i$ and iteratively moving through the sample sequence, decoding our model's prediction under the current dynamics, until the loss exceeds $\epsilon$. In this section, we describe how to vectorize this process efficiently over a batch of examples.

At a high level, we will compute the RNN cell update batch at each time step, but only update the parameters defining the linear region for instances that exceed the threshold. At each point in time, the model is within a linear segment defined by its starting point $h_{\tau_i}$ and its state velocity $\dot h_{\tau_i}$. 
Let the $H_{t}$ be a $B\times d$ matrix of starting points for the linear regions active at time $t$ for a batch of $B$ instances (i.e. $h_{\tau_i}$ if $t \in [\tau_i, \tau_i+\dti$). Likewise let ${M}_i$ be the slopes for the active linear regions. Let $\mathcal{M}$ be a $B$ dimensional, binary vector where the $b^{\mbox{th}}$ entry denotes whether the loss for the $b^{\mbox{th}}$ instance is greater than $\epsilon$ at the current step. With these definitions, we can write the batched update of $H_t$ and $M_t$ via a dynamic masking operation as follows:

\begin{align*}
    \hat{H}_{t}, \hat{\dot H}_{t} &= \mbox{RNNCell}\left([H_i,~ \dot{H}_i\dtiprev], v\left(\hat{X}(t)\right)\right)\\
    H_t &=       \mathcal{M}  \odot  H_t
        + (1-\mathcal{M}) \odot \hat{H}_t \\
     \dot H_t &=       \mathcal{M}  \odot  \dot H_t
        + (1-\mathcal{M}) \odot \hat{\dot H}_t \\
            &\qquad \qquad \text{where} \quad \mathcal{M} = \ell\left(x_t, \hat{x}(t)\right) < \epsilon
\end{align*}

And the matrix of hidden state vectors at $t$ can then be computed as:

\begin{equation}
    H(t) = H_t + (t-\tau_i)\dot H_t
\end{equation}

Since we must compute $\mathcal{M}$ dynamically as we progress over the time dimension, we cannot fuse the RNN cell updates over time, a common technique used to speed up teacher forcing. This, along with the additional cost of decoding $\hat H_t$ and evaluating prediction error, increases the runtime of a training step of our model by a factor of three over a baseline RNN. Unlike reference RNN implementations, however, our code is not optimized for speed. Furthermore, we are more interested in our model's performance during evaluation than during training, and thus are willing to tolerate this slowdown (which scales as a constant factor).

\subsection{Discussion of Bootstrapping}
\label{appendix:bootstrapping}
In early experiments, we noticed that our model had a difficult time transitioning from no jumpiness (selecting $\dt=1$ at every time step) to some jumpiness (selecting $\dt=2$ at rare intervals). Upon further investigation, we found that, having never needed to linearly extrapolate dynamics for more than one time step, our model was very bad at doing so early in training. In fact, the prediction errors at $\dt=2$ were consistently much higher than $\epsilon$. We considered two methods of solving this problem. One was to anneal $\epsilon$ starting from a large value to promote larger step sizes early in training. The other was to make the model to take an extra linear step 1\% of the time, regardless of the value of $\epsilon$. The second option gave the best and most stable results so we used it in our experiments.

\section{Additional experimental results}

\subsection{Visualizing the Lines Task}
\label{appendix:visualize_lines}

\begin{figure*}[b]
\centering
\includegraphics[width=\textwidth]{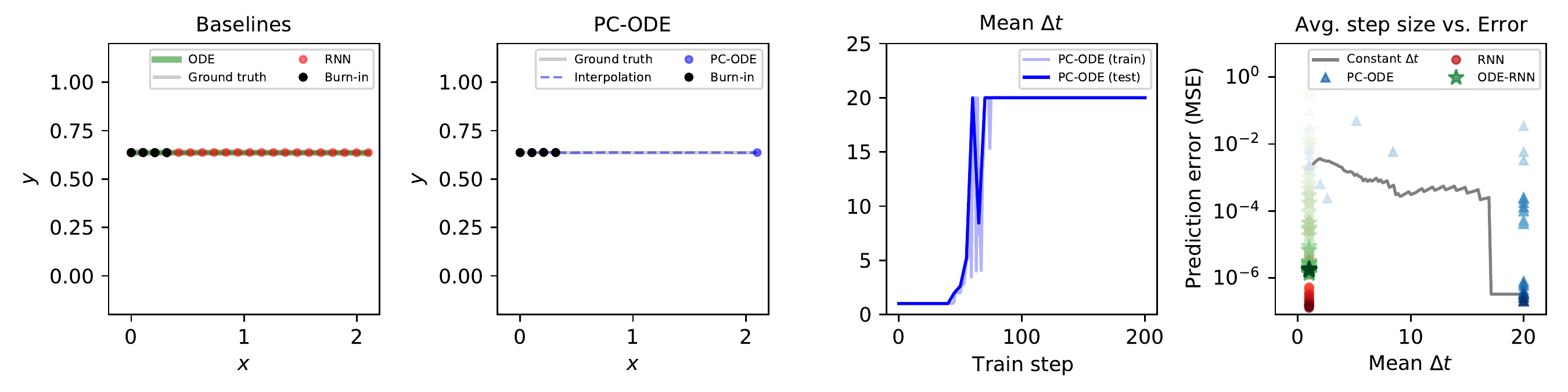}
\caption{A comparison between our model and the two baselines trained on the Lines task. \underline{Columns 1 and 2} show samples from the models. Here we see that the PC-ODE model can summarize an entire trajectory with a single linear function, whereas the baseline models cannot.  \underline{Columns 3 and 4} report key quantitative statistics of the two models. In particular, Column 3 shows how step size increases throughout training. Column 4 shows that the PC-ODE model is able to produce autoregressive samples of comparable quality to the baselines while using twenty times fewer cell updates.}
\label{fig:lines_summary}
\end{figure*}

\end{document}